\documentclass[a4paper, 10pt, conference]{ieeeconf}      
\usepackage{FG2025}

\FGfinalcopy 

\IEEEoverridecommandlockouts                              
\usepackage{graphicx}
\usepackage[alwaysadjust]{paralist}
\usepackage{amsmath} 
\usepackage{amssymb}  
\usepackage{booktabs}
\usepackage{multirow, bigdelim}
\usepackage{xcolor}
\newcommand{\etal}{\textit{et al.}}

\def\FGPaperID{193} 

\makeatletter
\let\NAT@parse\undefined
\makeatother
\usepackage[colorlinks]{hyperref}
\hypersetup{
    linkcolor=blue,
    filecolor=magenta,      
    urlcolor=cyan,
    citecolor=green    
    }

\title{\LARGE \bf
Using Cross-Domain Detection Loss to Infer Multi-Scale Information for Improved Tiny Head Tracking
}

\author{\parbox{16cm}{\centering
    {\large Jisu Kim$^1$, Alex Mattingly$^2$, Eung-Joo Lee$^3$ and Benjamin Riggan$^1$}\\
    {\normalsize
    $^1$ Department of Electrical and Computer Engineering, University of Nebraska-Lincoln, NE, USA\\
    $^2$ Department of Electrical and Computer Engineering, University of Maryland, MD, USA\\\vspace{-1.1px}
    $^3$ Department of Electrical and Computer Engineering, University of Arizona, AZ, USA}}
}

\usepackage{fancyhdr}

\begin{document}
\ifFGfinal
\thispagestyle{empty}
\pagestyle{empty}
\else
\author{Anonymous FG2025 submission\\ Paper ID \FGPaperID \\}
\pagestyle{plain}
\fi
\maketitle
\thispagestyle{fancy}

\begin{abstract}
Head detection and tracking are essential for downstream tasks, but current methods often require large computational budgets, which increase latencies and ties up resources (e.g., processors, memory, and bandwidth). To address this, we propose a framework to enhance tiny head detection and tracking by optimizing the balance between performance and efficiency. Our framework integrates (1) a cross-domain detection loss, (2) a multi-scale module, and (3) a small receptive field detection mechanism. These innovations enhance detection by bridging the gap between large and small detectors, capturing high-frequency details at multiple scales during training, and using filters with small receptive fields to detect tiny heads.  Evaluations on the CroHD and CrowdHuman datasets show improved Multiple Object Tracking Accuracy (MOTA) and mean Average Precision (mAP), demonstrating the effectiveness of our approach in crowded scenes.
\end{abstract}

\section{INTRODUCTION}

\begin{figure}
      \centering
      \includegraphics[width=1.0\columnwidth]{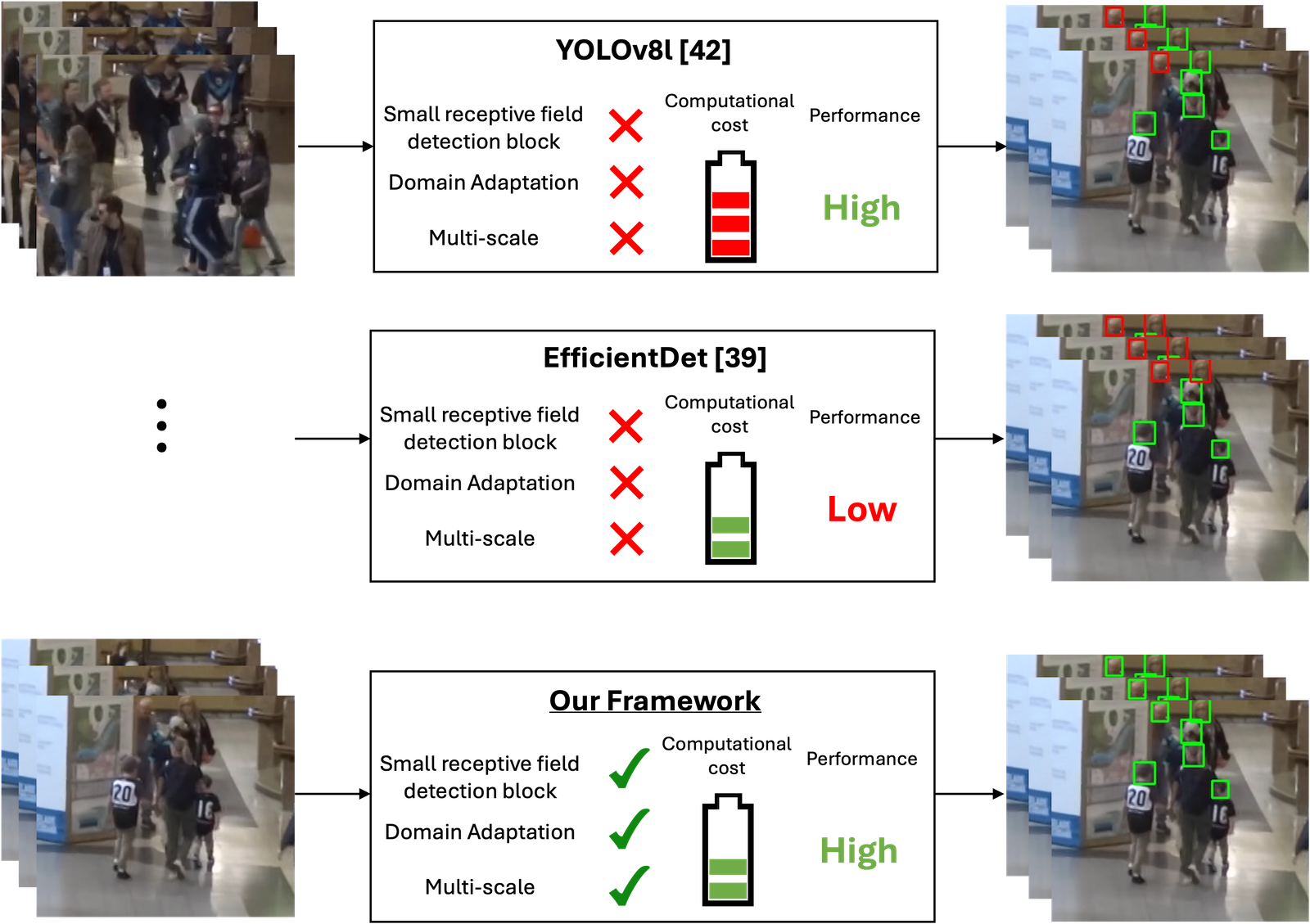}
      \caption{Comparison of head detection performance between the base models (top, middle) and our framework (bottom). The top model has high computational cost with good accuracy but lacks the small receptive field detection block, domain adaptation, and multi-scale processing. The middle model has lower computational cost but worse accuracy, and it also lacks these features. In contrast, our framework achieves better accuracy with lower computational cost by incorporating the small receptive field detection block, domain adaptation, and multi-scale processing.}
      \label{fig:highlevel}
\end{figure}

Head detection and tracking are critial prerequisite system components that are necessary for down-stream tasks, such as recognition, expression analysis, head pose estimation, gaze estimation, and person counting. However, the problem is that existing methods~\cite{bolme2010visual,li2024lightweight, zeng2024yolov8pd} often rely on large computational and energy resources (e.g., over 60 million parameters~\cite{li2024lightweight}, more than 100 Billion Floating-Point Operations Per Second (BFLOPs)~\cite{zeng2024yolov8pd}, or consuming upwards of 200 watts of power)~\cite{li2024lightweight}), which imposes severe limitations and bottlenecks for down-stream tasks. Moreover, these resource-intensive detection and tracking methods further hinder system scalability when considering operating in crowded environments (with greater than 50 persons in the field of view)~\cite{ge2021dense, qi2023pedestrian}. Thus, it is critical to achieve accurate detection and tracking across various resolutions, while also alleviating computational complexity (as shown in Figure~\ref{fig:highlevel}).  

Current approaches in head detection and tracking focus on achieving high accuracy, often at the expense of computational efficiency. Methods like Single Shot MultiBox Detector (SSD)~\cite{liu2016ssd}, Faster R-CNN~\cite{ren2015faster}, and RetinaNet~\cite{Lin2017} leverage deep neural networks to provide robust detection capabilities. However, these methods are computationally expensive, limiting their applicability in resource-constrained environments. Techniques such as Feature Pyramid Networks (FPNs)~\cite{lin2017feature} and EfficientDet~\cite{tan2020efficientdet} have been developed to address multi-scale detection, yet they still require significant computational resources. Although there are existing ``small'' or efficient variants of detection architectures, such as recen You Only Look Once (YOLO) models~\cite{YOLOv8,zeng2024yolov8pd,bochkovskiy2020yolov4}, these variants often exhibit significant reductions in performance.  Moreover, these existing methods often under perform when detecting tiny objects due to (1) down-sampling large images and (2) designing detectors to have relatively larger receptive fields.

In contrast, our framework aims to balance performance and computational efficiency by leveraging a cross-domain detection loss, a multi-scale module, and small receptive field detection mechanism. The cross-domain detection loss function bridges the gap between large and small detectors, while the multi-scale module captures additional high-frequency details early in the detection pipeline. By computing multi-scale information only during training, our proposed framework is able to infer multi-scale information without increasing computational complexity during inference—effectively enhancing the performance and efficiency trade-off. Additionally, the inclusion of a small receptive field detection mechanism enables predictions from early layers, which exploits local, fine-grained information to promote robust detection of tiny heads.

The key contributions of our framework include: 
\begin{compactitem}
    \item \textbf{A new cross-domain detection loss}---which promotes optimal performance from resource constrained models,
    \item \textbf{A new multi-scale module}---that combines features from different input resolutions \emph{during training,} enabling this information to be inferred by our domain adaptive detection framework without explicit computational overhead,
    \item \textbf{A newly integrated small receptive field detection mechanism}---which promotes the robust detection of relatively smaller (i.e., tiny) heads in crowded environments.
\end{compactitem}

Moreover, our proposed framework employs a tracking by detection approach, which like recent tracking algorithms (e.g., ByteTrack~\cite{zhang2022bytetrack} and Lee \etal~\cite{lee2022asilomar}), facilitates robust assignment of individual detection to consistent tracks. By enhancing the detection capabilities for tracking, our proposed framework achieves more accurate trajectory predictions using and improved tracking performance in real-time applications.

In summary, our framework addresses the limitations of existing head detection and tracking models by improving accuracy while also reducing computational complexity. By incorporating a cross-domain detection loss, a multi-scale module, and a small receptive field mechanism, the framework achieves a balanced trade-off between performance and efficiency. The approach shows consistent improvements on challenging datasets like Crowd of Heads Dataset (CroHD)~\cite{Sundararaman2021} and CrowdHuman~\cite{Shao2018}, proving effective for tiny head detection in crowded scenes and suitable for real-time use.

\addtolength{\textheight}{-3cm}   

\section{RELATED WORK}
In this section, existing approaches for head detection, head tracking, domain adaptation, and integrating multi-scale information are reviewed and contrasted with the proposed methodology.

\subsection{Head Detection}
Prominent detection frameworks like SSD~\cite{liu2016ssd} and Faster R-CNN~\cite{ren2015faster} balance speed and accuracy. Pyramid structures, such as FPN~\cite{lin2017feature} and RetinaFace~\cite{deng2020retinaface}, integrate features across layers for robust detection. Lightweight models like BlazeFace~\cite{bazarevsky2019blazeface} are efficient for real-time applications. Efforts like Zhu \etal~\cite{zhu2020tinaface} and Chen \etal~\cite{chen2021improving} enhance detection under challenging conditions using feature enhancement and adversarial learning.


Our approach enhances tiny head detection by integrating multi-scale training and cross-domain loss to improve accuracy in occluded and varied lighting scenarios.

\subsection{Head Tracking}
Tracking-by-detection generates bounding boxes per frame and associates them to create trajectories~\cite{Andriluka2008}. This technique is effective with deep learning-based object detection solutions~\cite{Ahmed2021}. Kalman filters~\cite{kalman1960new}, particle filters~\cite{arulampalam2002tutorial}, and correlation filters like MOSSE~\cite{bolme2010visual} provide robust tracking. Recent deep learning advances, such as SiamFC~\cite{bertinetto2016fully} and DCF-CSR~\cite{lukezic2017discriminative}, enhance tracking accuracy. Multi-object tracking algorithms, like SORT~\cite{bewley2016simple} and DeepSORT~\cite{wojke2017simple}, manage crowded scenes effectively. Methods like LDES~\cite{bhat2018unveiling} improve robustness under occlusions and varying lighting. YOLOv8~\cite{YOLOv8} offers high accuracy and speed for real-time applications, handling different resolutions and object sizes, even in crowded scenes. 

YOLOv8 employs three key loss functions for object detection: Complete Intersection over Union (CIoU)~\cite{zheng2020distance} loss, Distributed Focal Loss (DFL)~\cite{li2020generalized}, and Binary Cross-Entropy (BCE)~\cite{goodfellow2016deep} loss. These losses are defined as follows

\begin{equation}
L_{\text{CIoU}} = 1 - \text{IoU} + \frac{\rho^2(b, b_{gt})}{c^2} + \alpha v,
\end{equation}
\begin{equation}
L_{DFL} = -((\hat{y}_{i+1} - y) \log(p_{i}) + (y - \hat{y}_{i}) \log(p_{i+1})),
\end{equation}
\begin{equation}
L_{\text{BCE}} = (1 - t) \log(1 - \hat{t}) - t \log(\hat{t}),
\end{equation}
where $L_{\text{CIoU}}$ improves bounding box regression by considering overlap, center distance, and aspect ratio consistency, $L_{DFL}$ optimizes the distribution of bounding box boundaries, $L_{\text{BCE}}$ measures the difference between predicted and ground truth classification probabilities.

A detailed explanation of these loss functions, including their domain-adapted versions, can be found in Section~\ref{subsec:Domain Adaptation}. Extending from these losses, we establish the new cross-domain detection loss for training our framework.


\subsection{Domain Adaptation}
Domain adaptation improves model performance when training and testing data come from different distributions. Ganin \etal~\cite{Ganin2015} proposed DANN, using a gradient reversal layer to reduce domain discrepancy. Tzeng \etal~\cite{Tzeng2014} introduced DDC, minimizing domain differences with adaptation layers and a domain confusion loss, while Tzeng \etal~\cite{Tzeng2017} proposed ADDA, aligning feature distributions via adversarial learning. Hoffman \etal~\cite{Hoffman2018} leveraged GANs for domain adaptation, generating indistinguishable images between domains. French \etal~\cite{French2018} introduced self-ensembling, where a student model matches a teacher model's predictions under different augmentations. Chen \etal~\cite{Chen2020} proposed a Bi-directional Adaptation Network, adapting both source and target domains simultaneously. Saito \etal~\cite{Saito2018} used adversarial dropout regularization to align feature distributions.

Unlike these methods, our framework combines the strengths of large and small models through a cross-domain detection loss. This approach leverages the efficiency of the small model and the accuracy of the large model, enabling robust performance across diverse datasets without excessive computational resources.

\subsection{Multi-Scale Modules}
Multi-scale modules are crucial for detecting objects of varying sizes and resolutions. Lin \etal~\cite{lin2017feature} proposed the Feature Pyramid Network (FPN), which builds feature pyramids inside convolutional networks to detect objects at different scales by combining low-resolution, semantically strong features with high-resolution, semantically weak features. Lowe's SIFT~\cite{lowe2004distinctive} algorithm, initially used for image matching, detects and describes local features robust to scale changes, inspiring many scale-invariant detection methods. Recent advances in deep learning, such as multi-scale CNNs by He \etal~\cite{he2016deep}, have significantly improved detection accuracy.


YOLOv4~\cite{bochkovskiy2020yolov4} and EfficientDet~\cite{tan2020efficientdet} utilize FPN and BiFPN structures for efficient multi-scale detection. PANet~\cite{liu2018path} and Zhao \etal~\cite{zhao2017pyramid} improve information flow and feature integration across scales. YOLOv8~\cite{YOLOv8} combines features from multiple resolution branches to boost accuracy.

Our framework integrates a new multi-scale module that processes input images at multiple scales from the beginning of the detection pipeline, enhancing the detection of objects of various sizes while reducing computational load during inference.

\section{METHODOLOGY}

\begin{figure*}
      \centering
      \includegraphics[width=1.0\textwidth]{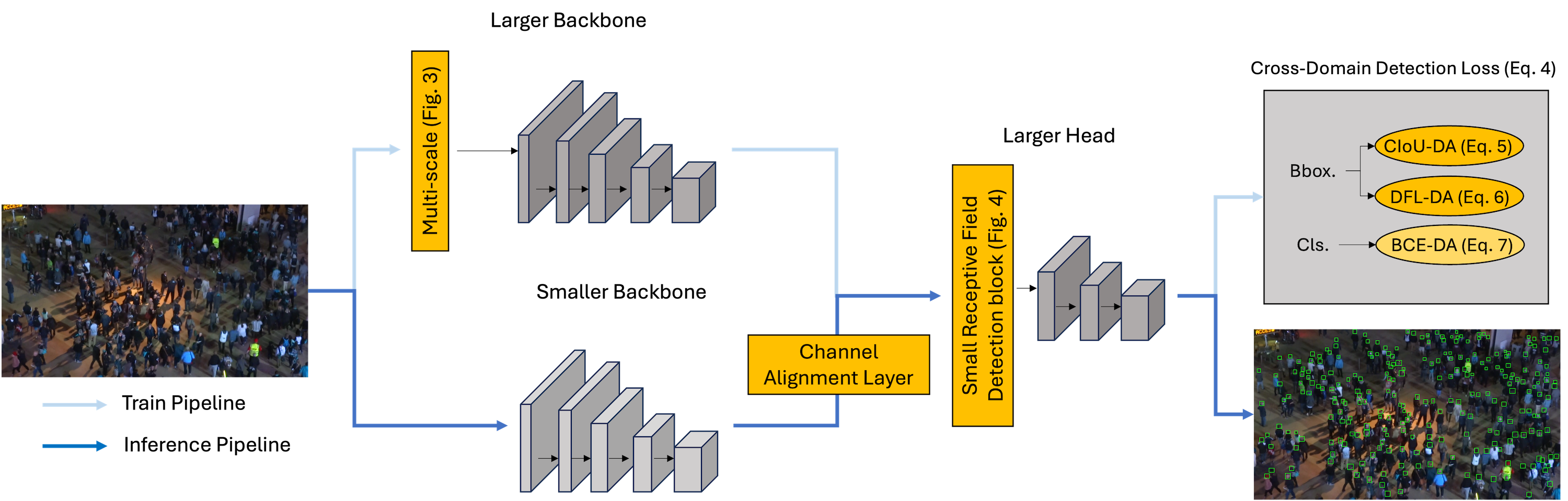}
      \caption{\textbf{Framework Overview}: In the training pipeline (both light and dark blue arrows), the multi-scale approach is applied to strengthen detection and tracking across various object sizes, while training is conducted using a large backbone and detection head. Moreover, the cross-domain detection loss is utilized during training to help improve the discriminability of features extracted from a more compact backbone architecture (details provided in Section~\ref{subsec:Domain Adaptation}). In the inference pipeline (dark blue arrows), only the small/compact backbone is used to reduce computational cost and detection is performed using large head via the channel alignment layer.}
      \label{fig:detail figure}
\end{figure*}

In this section, the technical design and implementation of the proposed framework (Figure~\ref{fig:detail figure}) is described in detail. While the proposed framework components are general and can be applied to many detection architectures, we primarily consider these components using YOLOv8 as an example architecture.

To enhance the performance, the new cross-domain detection loss, multi-scale module, and small receptive field detection block are incorporated. The cross-domain detection loss promotes improved detection/tracking performance without increasing computation complexity during inference. The multi-scale module improves detection accuracy by allowing the model to process input images at multiple scales, capturing objects of various sizes more effectively. It is important to emphasize that this multi-scale processing is applied only during training, ensuring no additional computational overhead during inference.

By integrating these techniques, our proposed model aims to achieve superior tradeoffs between detection/tracking performance and computational efficiency, significantly reducing both FLOPs and execution time while maintaining high accuracy.

\subsection{Cross-Domain Detection Loss}\label{subsec:Domain Adaptation}


In this work, we consider each domain---namely rich- and poor- quality---to be associated with a model's capacity to be sufficiently discriminative.
The rich-quality domain refers to predictions from a relatively higher-capacity model (e.g., YOLOv8-large), while the poor-quality domain corresponds to those from a lighter weight model (e.g., YOLOv8-small). Both models are trained on the same imagery, and the proposed cross-domain detection loss guides the lightweight model to produce outputs that are representationally similar to and consistent with more complex and higher performant models, thereby improving performance without increasing inference cost.

The proposed cross-domain detection loss combines a modified bounding box loss—comprising both the CIoU (Complete Intersection over Union)~\cite{zheng2020distance} loss (Equation~\ref{eq:CIoU_loss}) and DFL (Discrete Focal Loss)~\cite{li2020generalized} (Equation~\ref{eq:DFL})—and a modified classification loss—specifically BCE (Binary Cross-Entropy)~\cite{goodfellow2016deep} (Equation~\ref{eq:BCE_loss}).  These losses are modified in domain adaptive manner, where the proposed losses promoted improved compatibility between relatively high performing detection architectures and high efficient ones. Mathematically, the cross-domain detection loss is defined as
\begin{equation}
    \begin{split}
         L_{\text{CDDL}} = \lambda_1 \cdot L_{\text{CIoU\_DA}} + \lambda_2 \cdot L_{\text{DFL\_DA}} + \lambda_3 \cdot L_{\text{BCE\_DA}}, \label{eq:CDDL}   
    \end{split}
\end{equation}
where $L_{\text{CIoU\_DA}}$, $L_{\text{DFL\_DA}}$, $L_{\text{BCE\_DA}}$ are the domain-adapted losses for CIoU, DFL, and BCE, respectively, and $\lambda_{1}$, $\lambda_{2}$, $\lambda_{3}$ are the corresponding weighting factors for each of these losses. The weighting factors $\lambda_{1}$, $\lambda_{2}$, $\lambda_{3}$ are set to the same values as those used in the original YOLOv8~\cite{YOLOv8}.

The domain-adapted CIoU loss is defined as
\begin{equation}
\begin{split}
    L_{\text{CIoU\_DA}} = 2 - \text{IoU}(R, B) - \text{IoU}(P, B) + \frac{p_{\text{sum}}^2}{d^2} + 2\alpha\upsilon, 
\end{split}
\label{eq:CIoU_loss}
\end{equation}
where $\text{IoU}(R, B)$ and $\text{IoU}(P, B)$ represent the Intersection over Union between the predicted bounding boxes from rich-quality (R) and poor-quality (P) data and the ground truth box $B$, respectively. 
Here, $\text{IoU}(R, B)$ is an abbreviated form of $\text{IoU}(\text{bbox}(x_R, \theta_R), B)$, where $\text{bbox}(x_R, \theta_R)$ refers to the bounding box predicted from rich-quality (R) data. Similarly, $\text{IoU}(P, B)$ is an abbreviated form of $\text{IoU}(\text{bbox}(x_P, \theta_R), B)$, where $\text{bbox}(x_P, \theta_R)$ represents the bounding box predicted from poor-quality (P) data using the rich-quality model's parameters $\theta_R$. The Intersection over Union (IoU) is a metric that calculates the overlap between the predicted bounding box and the ground truth box $B$. The Intersection over Union (IoU) is defined as the ratio of the area of overlap between the two boxes to the area of their union. In this context, $\text{IoU}(R, B)$ measures how well the rich-quality prediction aligns with the ground truth, and $\text{IoU}(P, B)$ measures how well the poor-quality prediction aligns when using the rich-quality model's parameters.
The term $\frac{p_{\text{sum}}^2}{d^2}$ represents the normalized sum of the squared Euclidean distances between the predicted bounding boxes' center points and the ground truth center. Specifically, $p_{\text{sum}}^2 = p^2(R, b_{gt}) + p^2(P, b_{gt})$, where $p(R, b_{gt})$ and $p(P, b_{gt})$ are the Euclidean distances between the center points of the predicted bounding boxes from rich-quality and poor-quality data and the ground truth bounding box $b_{gt}$. The value $d$ is the diagonal length of the smallest enclosing box covering both the predicted and ground truth boxes. Finally, $2\alpha\upsilon$ represents the aspect ratio consistency term, which helps improve the bounding box's shape alignment.

The domain-adapted DFL is defined as
\begin{equation}
\begin{split}
    L_{\text{DFL\_DA}} = - (\hat{y}_{R,i+1} - y)(\log s_{R,i} + \log s_{P,i}) \\ - (y - \hat{y}_{R,i})(\log s_{R,i+1} + \log s_{P,i+1}), \label{eq:DFL}
\end{split}
\end{equation}
where $\hat{y}_{R,i}$ and $\hat{y}_{R,i+1}$ represent the predicted bounding box values at index $i$ and $i+1$ from the rich-quality (R) data, and $y$ is the ground truth bounding box value. 
The terms $s_{R,i}$ and $s_{P,i}$ are shorthand notations for the softmax values of the regression predictions. Specifically, $s_{R,i} = \text{softmax}(\text{reg}(z_R; \phi_R))_i$ corresponds to the softmax value at index $i$ for the rich-quality data, and $s_{P,i} = \text{softmax}(\text{reg}(z_P; \phi_R))_i$ corresponds to the softmax value at index $i$ for the poor-quality (P) data, both using the same model parameters $\phi_R$. 
Similarly, $s_{R,i+1}$ and $s_{P,i+1}$ represent the softmax values at index $i+1$ for the rich-quality and poor-quality data, respectively.

In this context, $\text{reg}(\cdot)$ refers to the bounding box regression function, which predicts the bounding box coordinates for an object based on the input features $z_R$ and $z_P$, and the model parameters $\phi_R$. The output of this regression function yields bounding box predictions that enable the loss to be computed.

The loss is calculated as a weighted sum of the logarithms of the softmax values, where the weights are determined by the differences between the predicted bounding box values $\hat{y}$ and the ground truth value $y$. Specifically, $(\hat{y}_{R,i+1} - y)$ and $(y - \hat{y}_{R,i})$ are the weights applied to the logarithms of the softmax values for both rich-quality and poor-quality data. The loss includes contributions from both $s_{R,i}$ and $s_{P,i}$, as well as from $s_{R,i+1}$ and $s_{P,i+1}$, reflecting the influence of both data qualities in the calculation.

The domain-adapted BCE loss is defined as
\begin{equation}
\begin{split}
    L_{\text{BCE\_DA}} = (1 - t) \log \left( (1 - c_R)(1 - c_P) \right) \\ + t \log (c_R c_P), \label{eq:BCE_loss}
\end{split}
\end{equation}
where $t$ represents the ground truth value. The terms $c_R = \text{cls}(u_R; \psi_R)$ and $c_P = \text{cls}(u_P; \psi_R)$ correspond to the classification probabilities predicted from the rich-quality (R) and poor-quality (P) data, respectively, using the same model parameters $\psi_R$. The term $\text{cls}(\cdot)$ refers to the classification function that outputs the probability of the predicted class being correct based on the input features $u_R$ or $u_P$ and the model parameters $\psi_R$. The first term, $(1 - t) \log \left( (1 - c_R)(1 - c_P) \right)$, penalizes incorrect predictions when the ground truth value $t$ is 0, by taking the logarithm of the product of the classification probabilities that predict the negative class (i.e., $1 - c_R$ and $1 - c_P$) for both rich-quality and poor-quality data. If the model predicts a high probability for the positive class when $t = 0$, the penalty increases, encouraging the model to predict low probabilities for the positive class when the ground truth is negative. The second term, $t \log (c_R c_P)$, rewards correct predictions when the ground truth value $t$ is 1, by taking the logarithm of the product of the classification probabilities from both data qualities. This loss function reflects the combined contributions of both rich-quality and poor-quality data in calculating the BCE loss, ensuring the model accounts for both data distributions.

\subsection{Multi-Scale Module}\label{subsec:Multiscale}
\begin{figure}
      \centering
      \includegraphics[width=1.0\columnwidth]{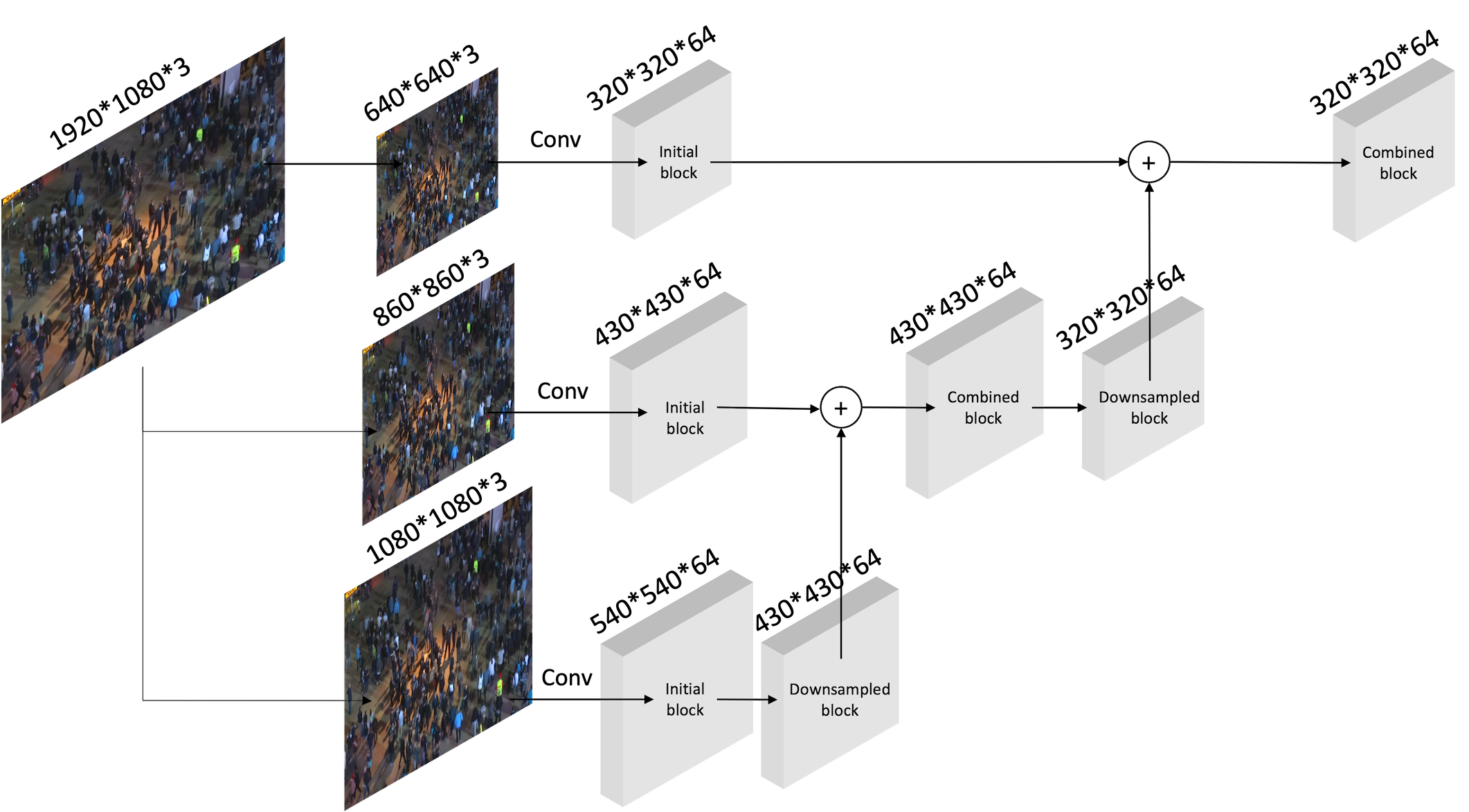}
      \caption{\textbf{Multi-scale module Overview}: The input image is downsampled multiple times to process images at multiple scales: 640×640, 860×860, and 1080×1080. Each branch undergoes convolution operations, and the resulting feature maps are merged through a series of downsampling and concatenation steps. This ensures that features at various scales are effectively captured and integrated to enhance the detection of different sized objects.}
      \label{fig:multiscale}
\end{figure}

Based on our experiments on the pre-trained models of the base model, we suspect that the first layer is not capable of properly extracting spatial information for tiny objects that degrades the model performance. Thus, we propose a new multi-scale module. Fig.~\ref{fig:multiscale} illustrates the multi-scale module process employed in our framework. The input image is initially downsampled into three branches of different sizes to capture features at multiple scales. Each branch undergoes a convolution operation with a stride of 2, resulting in the size of each block being halved and the depth increasing. This process creates three initial blocks of progressively smaller sizes.

The first merging step involves downsampling the largest initial block to match the size of the middle block. These two blocks are then concatenated. In the subsequent step, the combined block is further downsampled to match the size of the smallest initial block. The final concatenation merges this downsampled combined block with the smallest initial block, resulting in the ultimate combined block. This multi-scale module ensures that features at various scales are effectively captured and integrated, enhancing the model's ability to detect objects of different sizes more accurately.

\subsection{Integated Small Receptive Field Detection}\label{subsec:p2}
\begin{figure}
      \centering
      \includegraphics[width=1.0\columnwidth]{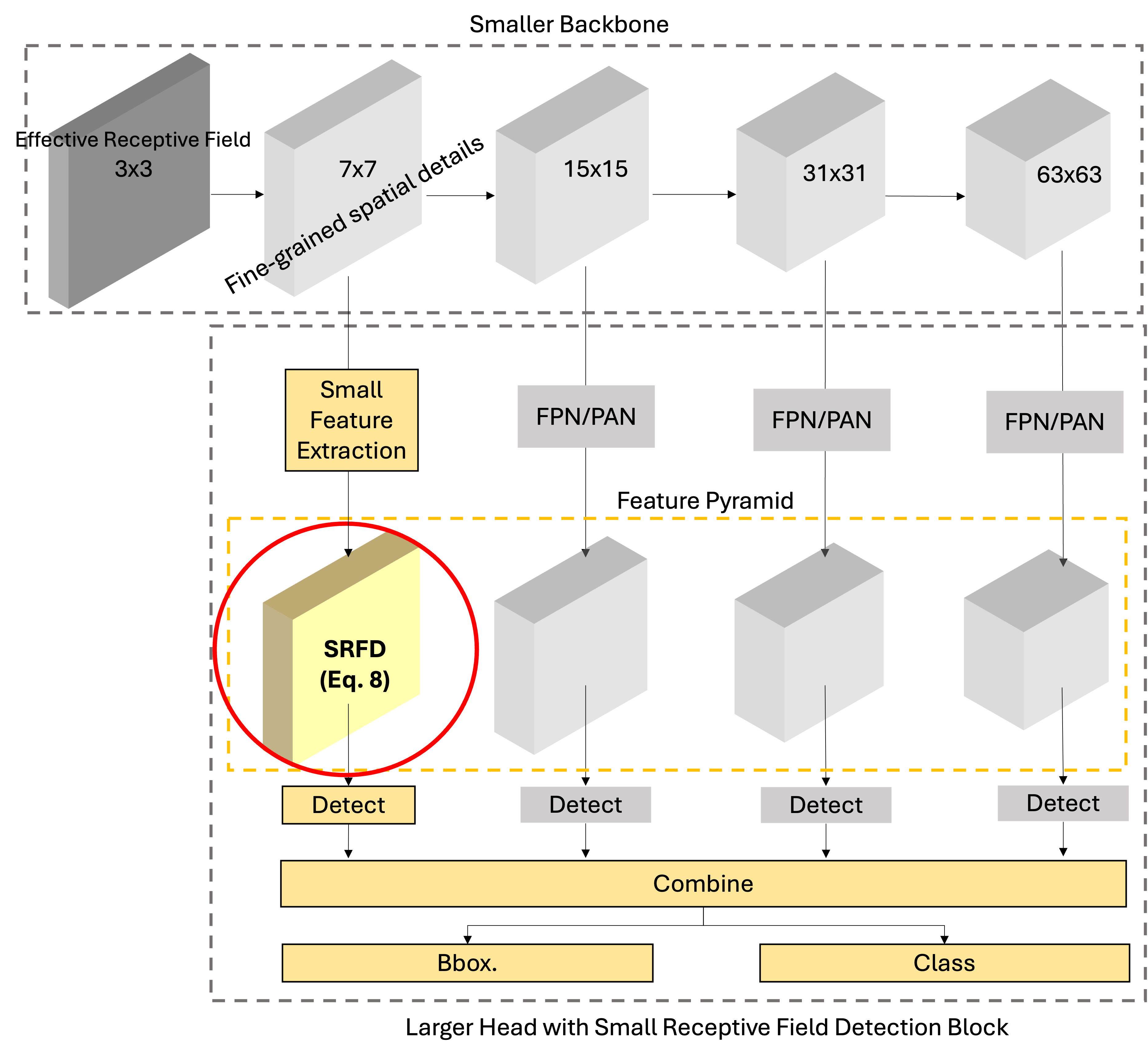}
      \caption{Backbone architecture and head with the addition of a \textbf{Small Receptive Field Detection (SRFD) block}. 
The orange components are added to operate on early features from the base model (gray sections). The SRFD block enhances the detection of small objects, resulting in improved performance for tiny object detection.}
      \label{fig:p2}
\end{figure}

In the base model, object detection relies on a three-layer Feature Pyramid, which struggles to detect tiny objects due to its relatively large receptive fields.

To address this, we introduce the Small Receptive Field Detection (SRFD) block, which enhances feature extraction at finer spatial resolutions. By adding this fourth layer to the Feature Pyramid, the model improves its ability to detect small objects, particularly in crowded or complex environments, as illustrated in Figure~\ref{fig:p2}. Mathematically, the SRFD block is formulated as
\begin{equation}
\text{SRFD block} = f_{3 \times 3} \Big( \phi \big( C \big( U(P_3) + f_{1 \times 1}(C_2), C_2 \big) \big) \Big), \label{eq:SRFD}
\end{equation}
where \( P_3 \) and \( C_2 \) are feature maps from different stages of the backbone. The function \( U(\cdot) \) upsamples \( P_3 \) to match the spatial resolution of \( C_2 \), while \( f_{1 \times 1}(\cdot) \) adjusts channel dimensions for effective fusion. The concatenation operation, \( C(\cdot) \), integrates upsampled high-level features with early-stage features, enhancing the preservation of fine-grained details. The bottleneck function, \( \phi(\cdot) \), reduces redundancy while maintaining key spatial information, allowing the network to focus on important object features. Finally, \( f_{3 \times 3}(\cdot) \) further refines and enhances small-object features before detection.

This additional layer strengthens multi-scale detection, leading to improved accuracy for tiny object detection and increased robustness against challenging scenarios.

\subsection{Tracking by Detection}
After identifying bounding boxes in each frame, we create trajectories by linking these boxes across frames using a tracking algorithm. This algorithm is based on the Generalized Intersection-Over-Union (GIoU) measure~\cite{rezatofighi2019generalized}. GIoU is an improvement of the IoU measure, as it takes into account the smallest rectangle that can contain both boxes. This helps in understanding the closeness of non-overlapping boxes since non-overlapping boxes have an IoU of 0.

Our algorithm works by iteratively selecting pairs of bounding boxes with the highest GIoU scores and updating the box in the next frame to match the one in the previous frame. Boxes that do not find a match are given a new unique ID. We also use a parameter, $\theta$, which is the minimum GIoU score required for a pair to be considered a match. In our experiments, we set $\theta$ to 0.

\section{EXPERIMENTS}

In this section, we describe the datasets, evaluation metrics, implementation details, and the results and analysis of our experiments.
\subsection{Dataset}
 We use two datasets for head tracking and detection. The first dataset is CroHD that video sequences from the Multi-Object Tracking (MOT) Benchmark Head Tracking 21 Challenge~\cite{Sundararaman2021}. These sequences contain images of populated pedestrian areas captured from an overhead camera. The entire dataset contains 2.2 million head bounding boxes in 11k images. Tracking information is also provided for training sequences. 

The other dataset is the CrowdHuman dataset~\cite{Shao2018} which is a benchmark for evaluating detectors in crowd scenarios, which is large, rich-annotated, and contains high diversity. It consists of 15k, 43.7k, and 5k images for training, validation, and testing, respectively. There are a total of 470k human instances from training and validation subsets and 22.6 persons per image, with various kinds of occlusions in the dataset. Each human instance is annotated with a head bounding-box, human visible-region bounding-box, and human full-body bounding-box.
\subsection{Evaluation}
The MOTA is Multiple Object Tracking Accuracy. This metric measures how well the tracker detects objects and predicts trajectories without taking precision into account. The metric takes into account three types of error~\cite{Bernardin2008}. The MOTA is computed as 
\begin{equation}
MOTA=1-\frac{\Sigma_t(FN_t+FP_t+IDSW_t)}{\Sigma_tGT_t}, \label{eq:MOTA}
\end{equation}
where $FN_t$ represents the number of false negatives, $FP_t$ denotes the number of false positives, $IDSW_t$ is the number of identity switches, and $GT_t$ indicates the number of ground truths.

\subsection{Results and Analysis}
\begin{figure}
      \centering
      \includegraphics[width=1.0\columnwidth]{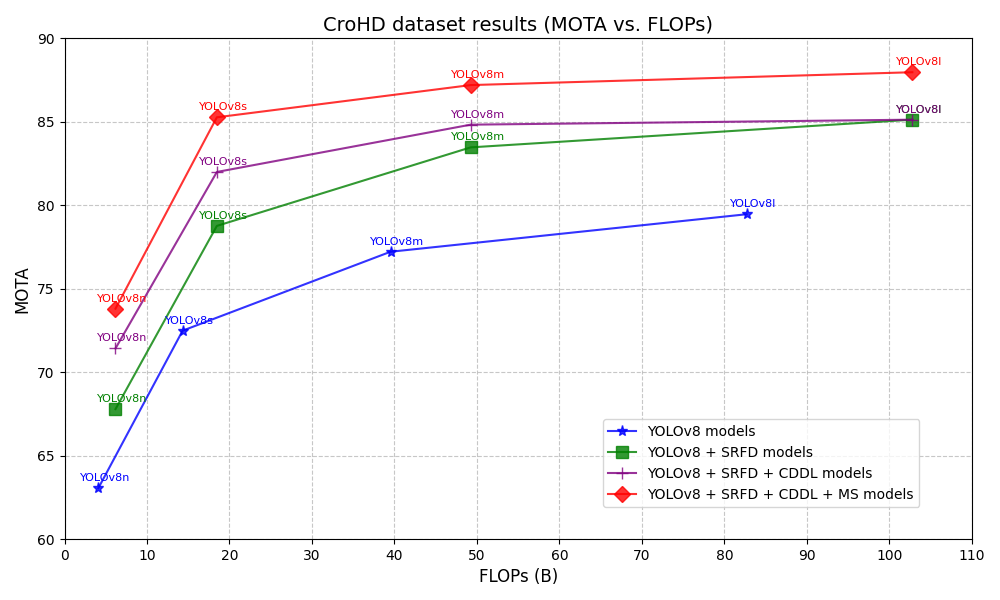}
      \caption{Comparison of MOTA vs. FLOPs on the CroHD dataset for sequences HT21-01 to HT21-04, consisting of a total of 5741 frames. The YOLOv8 models are represented by blue stars, YOLOv8 + SRFD models by green squares, YOLOv8 + SRFD + CDDL models by purple crosses, and YOLOv8 + SRFD + CDDL + MS models by red diamonds. The results show the trade-off between MOTA and computational complexity (FLOPs) for different configurations.}
      \label{fig:mota_FLOPs}
\end{figure}

Figure~\ref{fig:mota_FLOPs} illustrates the comparison of MOTA versus FLOPs on the CroHD dataset for sequences HT21-01 to HT21-04. The results are shown for various configurations of YOLOv8 models. Specifically, the YOLOv8 models are marked with blue stars, YOLOv8 + SRFD models with green squares, YOLOv8 + SRFD + CDDL models with purple crosses, and YOLOv8 + SRFD + CDDL + MS models with red diamonds.

Initially, the base YOLOv8 models (nano, small, medium, and large) demonstrate an increasing trend in MOTA as the FLOPs (inference parameters) increase. When SRFD is added to each model, there is a slight increase in FLOPs across all configurations, yet an improvement in MOTA of more than 5\% is observed. Furthermore, when both SRFD and CDDL are applied, the FLOPs remain unchanged compared to SRFD-only models, but MOTA further improves. Notably, applying CDDL to a large backbone and head does not yield additional gains compared to using only the large backbone.

Finally, with the integration of SRFD, CDDL, and MS, the training parameters increase, but the inference parameters remain constant, resulting in an overall trend of increased MOTA. This demonstrates the effectiveness of our proposed enhancements in balancing detection performance with computational efficiency.

\begin{table}
      \caption{MOTA performance, tracking only execution time, FLOPs, and MOTA per BFLOPs for our framework and other proposed solutions on the MOT leaderboard (sequences HT21-11 to HT21-15, consisting of a total of 5723 frames)}
      \label{tab:execution_time}
      \centering
      \resizebox{1.00\linewidth}{!}{
      \begin{tabular}{p{0.25mm}p{0.25mm}ccccc|c}
      \cline{3-8}\noalign{\smallskip}
      & & Tracker & MOTA$\uparrow$ & Time$\downarrow$ & Time(\/Frame)$\downarrow$ & FLOPs$\downarrow$ & \shortstack{MOTA \\ per BFLOPs}$\uparrow$ \\
      \noalign{\smallskip}\cline{3-8}\noalign{\smallskip}
      & & SORT~\cite{bewley2016simple} & 33.40 & 22.15s & 3.85ms & - & - \\
      & & Lee~\etal 2022~\cite{lee2022asilomar} & 46.80 & 215s & 37.40ms & 123.40B & 0.379 \\
      & & CTv0~\cite{Clustertracker2022} & 52.40 & 21.86s & 3.82ms & - & - \\
      & & HeadHunterT~\cite{Sundararaman2021} & 57.80 & 14307.5s & 2500ms & 22250B & 0.002 \\
      & & PHDTT~\cite{vo2022pedestrian} & 60.60 & 269.95s & 47.2ms & 419.81B & 0.144 \\
      & & FM\_OCSORT~\cite{cao2023observation} & 67.90 & 323.33s & 56.5ms & 502.80B & 0.135 \\ 
      & & YOLOv8s~\cite{YOLOv8} & 42.64 & 151.20s & 26.41ms & 14.32B & 2.977 \\ 
      & & YOLOv8l~\cite{YOLOv8} & 44.72 & 174.36s & 30.46ms & 82.70B & 0.540 \\ 
      
      \cline{3-8}
       \parbox[t]{2mm}{\multirow{4}{*}{\rotatebox[origin=c]{90}{\textbf{ours}}}} & \parbox[t]{2mm}{\multirow{4}{*}{\rotatebox[origin=c]{90}{$\overbrace{\hspace{0.4in}}$}}} & YOLOv8s + SRFD + CDDL & 46.60 & 157.44s & 27.51ms & 18.47B & 2.523 \\
       & & YOLOv8s + SRFD + CDDL + MS & 56.22 & 157.50s & 27.52ms & 18.47B & 3.043 \\ 
       & & YOLOv8l + SRFD & 47.00 & 187.60s & 32.78ms & 102.80B & 0.457 \\ 
       & & YOLOv8l + SRFD + MS & 57.13 & 187.49s & 32.76ms & 102.80B & 0.555 \\
      
      \noalign{\smallskip}\cline{3-8}
      \end{tabular}
      }
\end{table}

Table~\ref{tab:execution_time} presents the MOTA (Multiple Object Tracking Accuracy), total execution time, time per frame, FLOPs (Floating Point Operations), and MOTA per BFLOPs for our framework and other state-of-the-art tracking methods from the MOT leaderboard. The evaluation was conducted on sequences HT21-11 to HT21-15, consisting of a total of 5723 frames.

Among the evaluated methods, FM\_OCSORT~\cite{cao2023observation} achieves the highest MOTA of 67.90\%, but at a high computational cost of 502.80 BFLOPs and 56.5 seconds per frame. In contrast, our YOLOv8s + SRFD + CDDL + MS achieves a competitive MOTA of 56.22\% with only 18.47 BFLOPs and 17.35 seconds per frame, and a superior MOTA per BFLOPs of 3.043, outperforming FM\_OCSORT (0.135). This demonstrates our framework's ability to improve tracking accuracy without increasing computational complexity during inference.

The key advantage of our framework is its ability to enhance performance during training through CDDL, MS, and SRFD, while maintaining computational cost during inference. CDDL improves adaptability, MS ensures robustness across object sizes, and SRFD enhances small object detection, collectively improving accuracy without increasing FLOPs.

The lower execution time and FLOPs of our framework make it suitable for real-time applications, particularly in tracking tiny objects in crowded scenes. By integrating these modules, our framework improves accuracy while maintaining computational efficiency, making it ideal for resource-constrained environments.

For the results summarized in Table~\ref{tab:execution_time}, our frames were evaluated using the following devices: Intel i9-13900KF, NVIDIA Geforce RTX 4090. The results of other proposed solutions were obtained from the MOT leaderboard.


\begin{table}
      \caption{CrowdHuman dataset results. For our methods, multiple training resolutions were used, including 640x640, 840x840, and 1080x1080.}
      \label{tab:CrowdHuman dataset}
      \resizebox{\linewidth}{!}{
      \begin{tabular}{p{0.25mm}p{0.25mm}lcccc|c}
      \cline{3-8}\noalign{\smallskip}
      & & Method & Train Resolution(s) & mAP50(B)$\uparrow$ & mAP50-95(B)$\uparrow$ & FLOPs$\downarrow$ & \shortstack{mAP50-95(B) \\ per BFLOPs}$\uparrow$ \\
      \noalign{\smallskip}\cline{3-8}\noalign{\smallskip}
      
      & & SSD~\cite{liu2016ssd} & 300x300 & 0.631 & 0.388 & 2.25B & 0.172 \\
      & & Faster R-CNN~\cite{ren2015faster} & 800x800 & 0.657 & 0.402 & 48.00B & 0.008 \\
      & & RetinaNet~\cite{Lin2017} & 800x800 & 0.674 & 0.423 & 57.60B & 0.007 \\
      & & EfficientDet~\cite{tan2020efficientdet} & 512x512 & 0.682 & 0.428 & 12.58B & 0.034 \\
      & & DETR~\cite{Carion2020} & 800x800 & 0.668 & 0.405 & 96.00B & 0.004 \\
      & & YOLOv8n~\cite{YOLOv8} & 640x640 & 0.608 & 0.361 & 4.10B  & 0.088 \\
      & & YOLOv8s~\cite{YOLOv8} & 640x640 & 0.615 & 0.372 & 14.32B & 0.026\\
      & & YOLOv8m~\cite{YOLOv8} & 640x640 & 0.628 & 0.384 & 39.53B & 0.010 \\
      & & YOLOv8l~\cite{YOLOv8} & 640x640 & 0.642 & 0.395 & 82.70B & 0.005 \\
      \cline{3-8}
      \parbox[t]{2mm}{\multirow{4}{*}{\rotatebox[origin=c]{90}{\textbf{ours}}}}& \parbox[t]{2mm}{\multirow{4}{*}{\rotatebox[origin=c]{90}{$\overbrace{\hspace{0.4in}}$}}} & YOLOv8n + SRFD + CDDL + MS & multiple & 0.718 & 0.452 & 6.18B & 0.073 \\
      & & YOLOv8s + SRFD + CDDL + MS & multiple & 0.735 & 0.470 & 18.47B & 0.025 \\
      & & YOLOv8m + SRFD + CDDL + MS & multiple & 0.740 & 0.472 & 49.24B & 0.009 \\
      & & YOLOv8l + SRFD + MS & multiple & 0.742 & 0.474 & 102.80B & 0.004 \\
      \noalign{\smallskip}\cline{3-8}
      \end{tabular}
      }
\end{table}

\begin{figure*}
      \centering
      \includegraphics[width=1.0\textwidth]{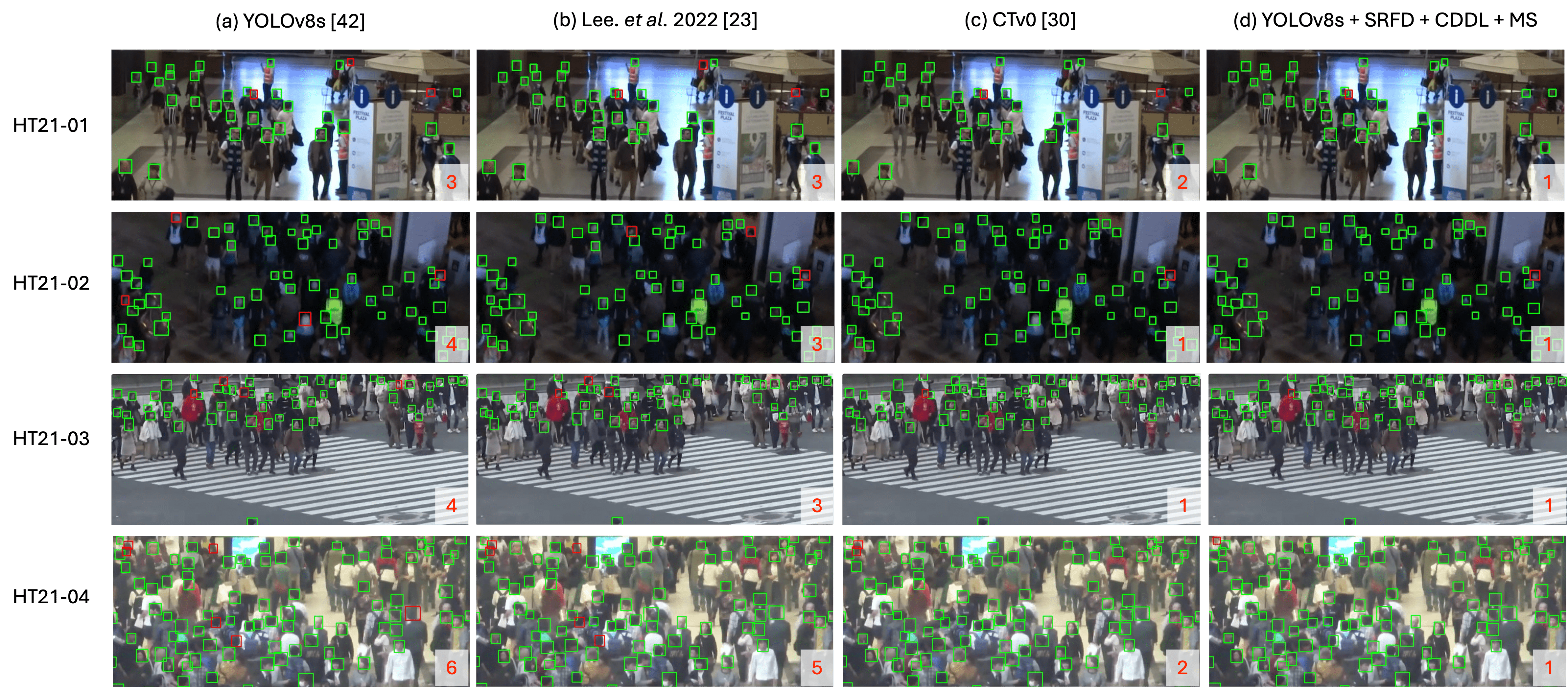}
      \caption{Comparative analysis of head detection results: (a) YOLOv8s~\cite{YOLOv8}, (b) Lee~\etal 2022~\cite{lee2022asilomar}, (c) CTv0~\cite{Clustertracker2022}, (d) YOLOv8s + SRFD + CDDL + MS. False negatives are indicated by red boxes. True positives are shown in green boxes. This comparative visualization highlights the differences in detection performance across various configurations, illustrating the strengths and weaknesses of each method.}
      \label{fig:bboxes results}
\end{figure*}

Table~\ref{tab:CrowdHuman dataset} compares various object detection models on the CrowdHuman dataset, including SSD, Faster R-CNN, RetinaNet, EfficientDet, and DETR. While DETR achieves a high mAP50 of 0.668 and mAP50-95 of 0.405, it requires 96.00 BFLOPs, resulting in a low mAP50-95 per BFLOPs of 0.004. EfficientDet achieves a competitive mAP50 of 0.682 and mAP50-95 of 0.428 with 12.58 BFLOPs, but its mAP50-95 per BFLOPs (0.034) is still lower than our framework.

Among the evaluated configurations, the proposed YOLOv8s + SRFD + CDDL + MS model achieves an mAP50 of 0.735 and an mAP50-95 of 0.470, outperforming the baseline YOLOv8s and state-of-the-art models like SSD (mAP50: 0.631) and Faster R-CNN (mAP50: 0.657). Notably, these improvements are achieved with only 18.47 BFLOPs, resulting in an mAP50-95 per BFLOPs of 0.025, which is significantly higher than EfficientDet (0.034) and other competing methods.

The key advantage of our framework is its ability to enhance performance during training through SRFD, CDDL, and MS, while maintaining computational cost during inference. SRFD improves small object detection, CDDL enhances adaptability across data distributions, and MS ensures robustness across object sizes, collectively improving accuracy without increasing FLOPs.

Furthermore, the proposed YOLOv8s + SRFD + CDDL + MS outperforms the larger YOLOv8l + SRFD model, which achieves slightly higher mAP50 (0.742) and mAP50-95 (0.474) but at a significantly higher computational cost (102.80 BFLOPs). Our framework achieves comparable accuracy with only 18.47 BFLOPs, making it highly efficient for real-time applications in resource-constrained environments.

Figure~\ref{fig:bboxes results} presents a comparative analysis of head detection results on the CroHD dataset. From left to right, the images show results from (a) YOLOv8s~\cite{YOLOv8}, (b) Lee~\etal 2022~\cite{lee2022asilomar}, (c) CTv0~\cite{Clustertracker2022}, and (d) YOLOv8s + SRFD + CDDL + MS. Red boxes indicate false negatives, and green boxes indicate true positives. The results demonstrate the progressive improvement in detection accuracy as SRFD, CDDL, and MS are integrated, with the number of red boxes decreasing significantly in (d). This visualization highlights the strengths of our proposed framework in detecting tiny heads in crowded scenes.

\section{CONCLUSIONS}
In this study, we presented an enhanced approach for tiny head detection and tracking, integrating cross-domain detection loss, multi-scale module, and the addition of the small receptive field detection block. Our framework combines the efficiency of the small backbone with the accuracy of the large detection head trained by cross-domain detection loss, ensuring robust performance across diverse datasets.

The multi-scale module, coupled with the small receptive field detection block, significantly improves the detection of small objects by leveraging features at different scales. This enhancement is particularly effective in real-world scenarios where object sizes can vary widely. Evaluations using the CroHD dataset from the MOT Challenge Head Tracking 21 and additional experiments on the CrowdHuman dataset demonstrate that our model achieves superior performance in terms of MOTA and mAP metrics, showcasing its ability to accurately detect and track tiny heads in crowded and complex environments.

Our experimental results indicate that the integration of cross-domain detection loss and multi-scale module, along with the small receptive field detection block, provides a balanced solution that meets the demands of real-time applications with limited computational resources. This work highlights the potential of combining advanced detection and tracking techniques to enhance performance in challenging conditions. Future work will focus on further optimizing the computational efficiency and exploring the application of these methods to other object detection and tracking tasks in various domains.

\section{ACKNOWLEDGMENTS}
This research is based upon work supported by DEVCOM Army Research Laboratory (ARL) under contract W911NF-21-2-0076. The views and conclusions contained herein are those of the authors and should not be interpreted as necessarily representing the official policies, either expressed or implied, of DEVCOM ARL or the U.S. Government. The U.S. Government is authorized to reproduce and distribute reprints for governmental purposes not withstanding any copyright annotation therein.
\clearpage
\section*{Ethical Impact Statement}

Our research focuses on detecting and tracking tiny heads in crowded environments. Since we use datasets that include human faces, there are potential ethical concerns related to privacy and identification. However, our work does not aim to recognize or identify individuals. The detected heads exhibit very few pixels across the face and lack discriminative facial details, which makes  identification extremely difficult and unlikely.

Additionally, we only use publicly available datasets (CroHD and CrowdHuman). These datasets are widely used in research and were created following ethical guidelines. We do not collect any new personal identifiable data, and we do not modify or enhance the resolution of the faces in the datasets. This ensures that our study does not increase privacy risks beyond what already exists publicly.

To further reduce ethical concerns, our research focuses on improving detection performance while maintaining efficiency, rather than using personal characteristics. We believe our work has positive impacts, such as better tracking in safety-critical areas (e.g., crowd management) without violating privacy. However, we acknowledge the importance of ethical AI development and encourage future researchers to continue discussing responsible use of AI in real-world applications.


{\small
\bibliographystyle{ieee}
\bibliography{egbib}
}

\end{document}